%% file: s4-arxiv.tex
\documentclass[letterpaper, twocolumn]{article}
\usepackage{geometry}
\usepackage{times}
\usepackage{courier}
\usepackage{graphicx}
\usepackage[hidelinks]{hyperref}
\usepackage{natbib}
\usepackage{caption}
\usepackage{authblk}
\usepackage[final]{fixme}

\usepackage{booktabs}
\usepackage{subcaption}
\usepackage{mathtools}
\usepackage{multirow}
\usepackage{algorithmicx, algorithm, algpseudocode}
\usepackage{amsmath,amssymb,amsthm}

\DeclarePairedDelimiterX{\infdivx}[2]{(}{)}{%
  #1\;\delimsize\|\;#2%
}

\newcommand{\argmax}{\mathop{\mathrm{arg\,max}{}}}

\newcommand{\cV}{{\mathcal{V}}}
\newcommand{\eat}[1]{\ignorespaces}

\renewenvironment{abstract}{\centerline{\bf
Abstract}\vspace{0.5ex}\begin{quote}\small}{\par\end{quote}\vskip 1ex}

\title{Self-supervised self-supervision by combining deep learning and probabilistic logic}
\date{}

\author[1]{Hunter~Lang}
\author[2]{Hoifung~Poon}

\affil[1]{MIT}
\affil[2]{Microsoft Research}
\affil[ ]{\small{\texttt{hjl@mit.edu, hoifung@microsoft.com}}}

\begin{document}

\maketitle

\input{sections/abstract}
\input{sections/intro}
\input{sections/dpl}
\input{sections/s4}
\input{sections/exp}
\input{sections/related}

%%%%%%%%%%%%%%%%%%%%%%%%%%%%%%%%%%%%%%%%%%%%%%%%%%%%%%%%%%%%%%%%%%%%%%%%%%%%%%%%%%
\section{Conclusion}

We present Self-Supervised Self-Supervision (S4), a general self-supervision framework that can automatically induce new self-supervision by extending deep probabilistic logic (DPL) with structure learning and active learning capabilities.
Our experiments on various natural language processing (NLP) tasks show that
compared to prior systems for task-specific self-supervision, such as Snorkel and DPL, S4 can obtain gain up to 20 absolute accuracy points with the same amount of supervision.
S4 only relies on humans to identify the most salient self-supervision for initialization and to verify proposed self-supervision, which tends to be the most effective use of human bandwidth.
While we focus on NLP tasks in this paper, our methods are general and can potentially be applied to other domains.
Future directions include: further investigation in combining structure learning and active learning; exploring more sophisticated self-supervision classes and proposal algorithms; applications to other domains.
\section{Acknowledgments}
We give warm thanks to Naoto Usuyama, Cliff Wong, and the rest of Project Hanover team for helpful discussions.

\bibliographystyle{apalike}
\bibliography{refs}
\clearpage
\input{sections/appendix}
\end{document}

%% file: sections/abstract.tex
\begin{abstract}
Labeling training examples at scale is a perennial challenge in machine learning. Self-supervision methods compensate for the lack of direct supervision by leveraging prior knowledge to automatically generate noisy labeled examples. Deep probabilistic logic (DPL) is a unifying framework for self-supervised learning that represents unknown labels as latent variables and incorporates diverse self-supervision using probabilistic logic to train a deep neural network end-to-end using variational EM.  While DPL is successful at combining pre-specified self-supervision, manually crafting self-supervision to attain high accuracy may still be tedious and challenging. In this paper, we propose Self-Supervised Self-Supervision (S4), which adds to DPL the capability to learn new self-supervision automatically. Starting from an initial ``seed,'' S4 iteratively uses the deep neural network to propose new self supervision. These are either added directly (a form of \emph{structured} self-training) or verified by a human expert (as in feature-based active learning). Experiments show that S4 is able to automatically propose accurate self-supervision and can often nearly match the accuracy of supervised methods with a tiny fraction of the human effort.
\end{abstract}

%% file: sections/intro.tex
\section{Introduction}
Machine learning has made great strides in enhancing model sophistication and learning efficacy, as exemplified by recent advances in deep learning \citep{lecun2015deep}. However, contemporary supervised learning techniques require a large amount of labeled data, which is expensive and time-consuming to produce. This problem is particularly acute in specialized domains like biomedicine, where crowdsourcing is difficult to apply.
Self-supervision has emerged as a promising paradigm to overcome the annotation bottleneck by automatically generating noisy training examples from unlabeled data. 
In particular, {\em task-specific self-supervision} converts prior knowledge into self-supervision templates for label generation,
as in distant supervision \citep{mintz2009distant}, data programming \citep{ratner2016data}, and joint inference \citep{poon2008joint}.

{\em Deep probabilistic logic (DPL)} is a unifying framework for self-supervision that combines deep learning with probabilistic logic \citep{wang2018deep}. It represents unknown labels as latent variables, and incorporates prior beliefs over labels and their dependencies as virtual evidence in a graphical model.
The marginal beliefs over the latent variables are used as probabilistic labels to train a deep neural network for the end prediction task. The trained neural network in turn provides belief updates to refine the graphical model parameters, and the process continues, using variational EM.

\citet{wang2018deep} show that DPL can effectively combine diverse sources of self-supervision in a coherent probabilistic framework and subsume supervised and semi-supervised learning as special cases. While promising, DPL and related approaches still require human experts to manually specify self-supervision.
This is particularly challenging for self-supervision techniques such as data programming and joint inference, which require domain expertise and extensive effort to identify the many relevant virtual evidences for attaining high accuracy in the end task. 
\setlength{\textfloatsep}{10pt}
\begin{figure*}[t]
\centering
\includegraphics[width=0.89\linewidth]{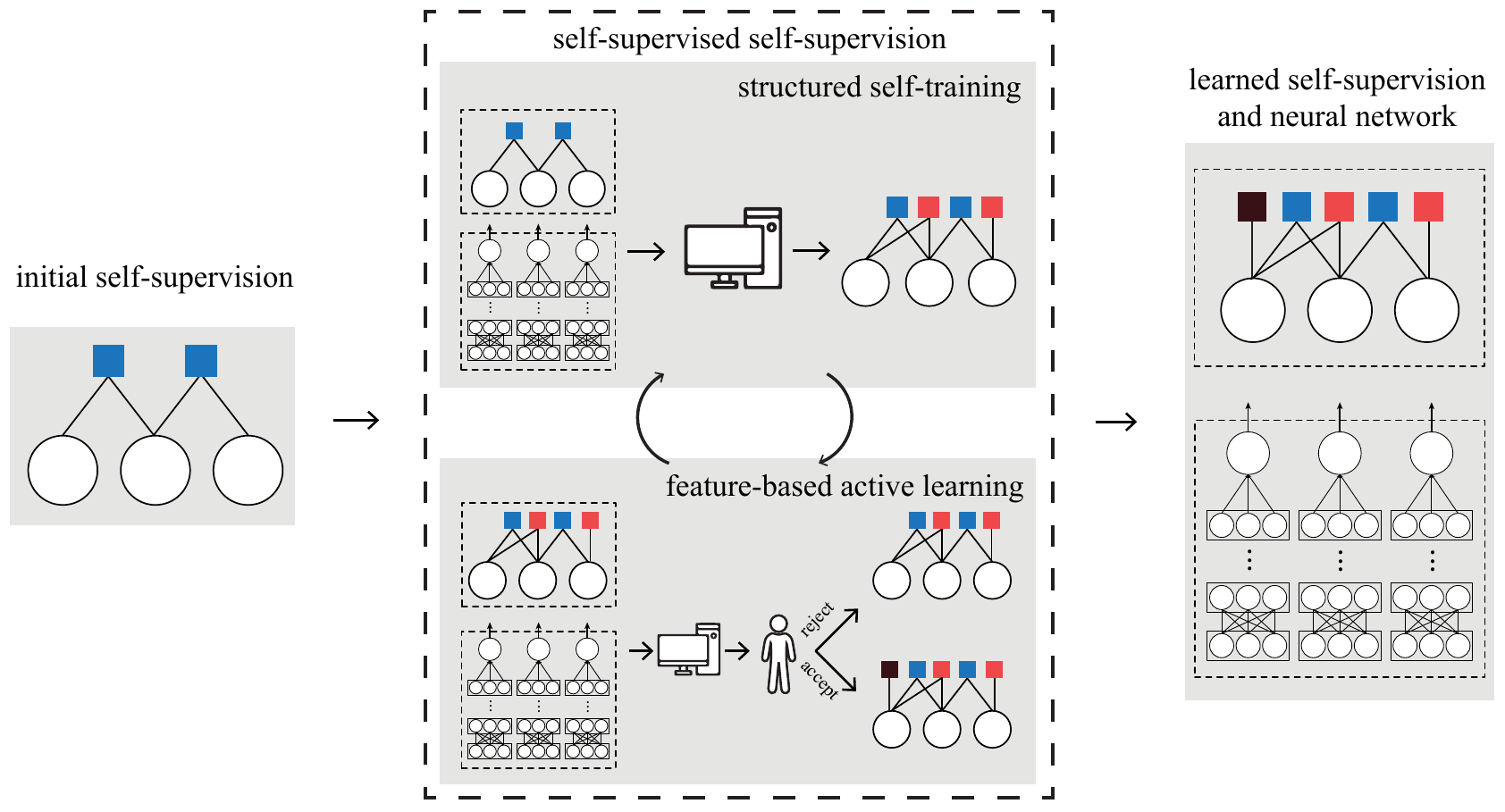}
\caption{Self-supervised self-supervision (S4): S4 builds on deep probabilistic logic and uses probabilistic logic to represent self-supervision for learning a deep neural network for the end prediction task. 
Starting from pre-specified self-supervision, S4 interleaves structure learning and active learning steps to introduce new self-supervision for training the neural network and refining the graphical model parameters. Self-supervision factors from initialization, structure learning, and active learning are shown in blue, red, and black, respectively. 
}
\label{fig:S4}
\end{figure*}

In this paper, we propose {\em self-supervised self-supervision (S4)} as a general framework for learning to add new self-supervision. In particular, we extend deep probabilistic logic (DPL) with structure learning and active learning capabilities (see Figure~\ref{fig:S4}). 
After running DPL using the pre-specified seed self-supervision,
S4 iteratively proposes new virtual evidence using the trained deep neural network and graphical model, and determines whether to add this evidence directly to the graphical model or ask a human expert to vet it. The former can be viewed as {\em structured self-training}, which generalizes self-training \citep[e.g.,][]{mcclosky2006effective} by adding not only individual labels but also arbitrary probabilistic factors over them.
The latter subsumes {\em feature-based active learning} \cite{druck2009active} with arbitrary features expressible using probabilistic logic. 
By combining the two in a unified framework, S4 can leverage both paradigms for generating new self-supervision and subsume many related approaches as special cases.

We use transformer-based models for the deep neural network in DPL and explore various self-supervision proposal mechanisms based on neural attention and label entropy. 
Our method can learn to propose both unary potential factors over individual labels and joint-inference factors over multiple labels.
We conducted experiments on various natural language processing (NLP) tasks to explore the potential of our method. We held out gold labels for evaluation only, and used them to simulate oracle self-supervision for initial self-supervision and active learning.
We find that S4 can substantially improve over the seed self-supervision by proposing new virtual evidence, and can match the accuracy of fully supervised systems with a fraction of human effort.

%% file: sections/dpl.tex
\section{Deep probabilistic logic}
Given a prediction task, let $\mathcal{X}$ and $\mathcal{Y}$ denote the sets of possible inputs and outputs, respectively. 
The goal is to train a prediction module $\Psi(x,y)$ that scores output $y$ given input $x$.
We assume that $\Psi(x,y)$ represents the conditional probability $P(y|x)$.
Let $X=(X_1,\cdots,X_N)$ denote a sequence of inputs and $Y=(Y_1,\cdots,Y_N)$ the corresponding outputs. 
If $Y$ is observed, $\Psi(x,y)$ can be learned using standard supervised learning.
In this paper, we consider the setting where $Y$ is unobserved, and $\Psi(x,y)$ is learned using self-supervision. 

The key idea of deep probabilistic logic (DPL) is to represent self-supervision as prior belief over the latent label variables Y and their interdependencies by combining probabilistic logic and deep learning \citep{wang2018deep}.
\citet{pearl1988probabilistic} first introduced {\em virtual evidence} to represent prior belief on the value of a random variable. Specifically, the prior belief on $Y$ can be represented by introducing a binary variable $v$ as a dependent of $Y$ such that $P(v=1|Y=y)$ is proportional to the belief of $Y=y$. 
The virtual evidence $v=1$ can be viewed as a reified variable representing the unary potential function $\Phi(y)\propto P(v=1|y)$. More generally, this can represent arbitrary potential functions $\Phi(X,Y)$ over the inputs and outputs, so as to model prior beliefs over arbitrary high-order factors. 
DPL uses Markov logic \cite{richardson&domingos06} to represent virtual evidences and uses a deep neural network as the prediction module $\Psi(x, y)$.
Let $\textbf{V}=\{v_1, \cdots, v_k\}$ be a set of pre-specified virtual evidences, with the corresponding potential functions being $(\Phi_1,\cdots,\Phi_k)$. We will use $K$ as shorthand for the event $\textbf{V}=\textbf{1}$ (i.e., $v_1=1, \ldots, v_k=1$). 
%The variables $v_k$ correspond to the available pieces of self-supervision and are always ``observed'' to be 1.
DPL defines a probability distribution over $K,X,Y$ by combining a factor graph with potentials $\Phi$ representing $P(K|Y,X)$ and the prediction module $P(Y|X)$:
\begin{equation}
  \label{eqn:dpl-factorization}
  P(K,Y|X)\propto \prod_v~\Phi_{v}(X, Y)\cdot\prod_i~\Psi(X_i, Y_i)
\end{equation}
%over the virtual evidence and the latent labels $Y$. 
Here, the potential functions $\Phi_v(X,Y)$ are represented by weighted first-order logic formulas (i.e.,~$\Phi_v(X,Y)=\exp(w_v f_v(X,Y))$, with $f_v(X,Y)$ being a binary feature represented by a first-order logical formula). 
DPL considers a Bayesian setting where each $w_v$ is drawn from a pre-specified prior distribution $w_v\sim P(w_v|\alpha_v)$. \fxnote{missing example with text-based factor; we told the reviewers we'd mention the running example earlier in the paper}
Fixed $w_v$ amounts to the special case when the prior is concentrated on the preset value. 
For uncertain $w_v$'s, DPL computes their maximum a posteriori (MAP) estimates.

Parameter learning in DPL maximizes the conditional log likelihood of virtual evidences $\log P(K|X)$, which can be done using variational EM. 
In the E-step, DPL computes a variational approximation $q(Y)$ for $P(Y|K,X)$, 
using loopy belief propagation \citep{murphy1999loopy} with current parameters $\Phi, \Psi$, by conducting message passing in $P(K,Y|X)$ iteratively. 
In the M-step, DPL treats $q(Y)$ as the probabilistic label distribution to train $\Phi$ and $\Psi$ via standard supervised learning.
For the prediction module $\Psi$, this reduces to standard deep learning, with the marginals $q_i(Y_i)=\mathbb{E}_{q(Y)}(Y_i)$ serving as probabilistic labels for $X_i$. 
For the supervision module, this reduces to standard parameter learning for log-linear models (i.e., learning non-fixed $w_v$'s), and can be solved using gradient descent, with the partial derivative for $w_v$ being
$\mathbb{E}_{\Phi(Y,X)}~[f_v(X,Y)] - \mathbb{E}_{q(Y)}~[f_v(X,Y)]$.
%For a tied weight, the partial derivative will sum over all features that originate from the same template.
The second expectation can be done by simple counting. The first expectation, on the other hand, requires probabilistic inference in the graphical model. But it can be computed using belief propagation, similar to the E-step, except that the messages are limited to factors in the supervision module (i.e., messages from $\Psi$ are not included)\footnote{In theory, to optimize a lower bound on the conditional log likelihood as in standard EM, E-steps and M-steps need to use different approximate inference algorithms \citep{domke2013learning}. However, using the same approximation for both has been shown to work well in practice \citep{verbeek2008scene}.}.

%% file: sections/s4.tex
\setlength{\textfloatsep}{10pt}
\begin{algorithm}[t]
\begin{algorithmic}
\caption{Self-Supervised Self-Supervision (S4)}\label{alg:S4}
\State \textbf{Input:} Seed virtual evidences $I$, deep neural network $\Psi$, inputs $X=(X_1,\ldots,X_N)$, unobserved outputs $Y=(Y_1,\ldots,Y_N)$, human query budget $T$.
\State \textbf{Output:} Learned prediction module $\Psi$ and virtual evidences $K = \{(f_v(X,Y), w_v):v\}$.
\State \textbf{Initialize:} $K=I$;  $Q=\emptyset$; $i  = 0$.
%\For{$i = 1..M$; $i<M$; $i$\small{++}}
\For{$i = 1\ldots M$}
\While{Structured Self-Training not converged}
    \State $\Psi, K \gets \mbox{\texttt{DPL-Learn}}(K, \Psi, X, Y)$
    \State $v=\texttt{PropSST}(K, \Psi, X, Y)$; 
    \State $K\gets K\cup v$; 
\EndWhile
\If{$|Q|<T$}
\State $v=\mbox{\texttt{PropFAL}}(K, \Psi, X, Y, Q)$;
\State $Q\gets Q \cup v$;
\State {\bf if} $\mbox{\texttt{Human-Accept}}(v)$ {\bf then} $K \gets K \cup v$;
\EndIf
\EndFor
\end{algorithmic}
\end{algorithm}

\section{Self-supervised self supervision}
\label{sec:s4}

%% overview
In this section, we present the {\em self-supervised self supervision (S4)} framework, which extends deep probabilistic logic (DPL) with the capability to learn new self-supervision.
Let~$\cV=\{(f_v,w_v,\alpha_v):v\}$~be the set of all candidate virtual evidences, where $f_v(X,Y)$ is a first-order logical formula, $w_v$ is the weight, and $\alpha_v$ is the weight prior (for non-fixed $w_v$).
Let $K$ be the set of virtual evidences maintained by the algorithm, initialized by the seed $I$. 
The key idea of S4 is to interleave structure learning and active learning to iteratively propose new virtual evidence $v\in \cV$ to augment $K$ (Figure~\ref{fig:S4}). 
Self-training is a special case where candidate virtual evidences are individual label assignments (i.e., $f_v=I[y_v=l_v]$).
S4 can thus be viewed as conducting {\em structured self-training (SST)} by generalizing self-training to admit arbitrary Markov logic formulas as virtual evidence.

S4 can also be viewed as conducting {\em structure learning} in the factor graph that specifies the virtual evidence. Structure learning has been studied intensively in the graphical model literature \citep{koller-struc-lrn}. It is also known as feature selection or feature induction in general machine learning literature \citep{feat-sel}.
Here, we are introducing structured factors for self-supervision, rather than as feature templates to be used during training. 
Another key difference from standard structure learning is the deep neural network, which provides an alternative view from the virtual evidence space and enables multi-view learning in DPL. The neural network can also help identify candidate virtual evidences, e.g., via neural attention.

%\vspace{-8pt}
In data programming and many other prior methods, human experts need to pre-specify all self-supervision upfront. While it is easy to generate a small seed by identifying the most salient self-supervision, this effort can quickly become tedious and more challenging as the experts are required to enumerate the less salient templates. On the other hand, given a candidate, it's generally much easier for experts to validate it. {\em This suggests that for the best utilization of human bandwidth, we should focus on leveraging them to produce the initial self-supervision and verify candidate self-supervision.}
Consequently, in addition to structured self-training (SST), S4 incorporates {\em feature-based active learning (FAL)} (i.e., active learning of self-supervision). When SST converges, S4 will switch to the active learning mode by proposing a candidate virtual evidence for human verification (i.e., labeling a feature rather than an instance in standard active learning). Intuitively, in FAL we are proposing virtual evidences for which the labels of the corresponding instances are still uncertain. If the human expert can provide definitive supervision on the label, the information gain will be large.
By contrast, in SST, we favor virtual evidences with skewed posterior label assignments for their corresponding instances, as they can potentially amplify the signal.

Algorithm~\ref{alg:S4} describes the S4 algorithm. S4 first conducts DPL using the initial self-supervision $I$, then interleaves structured self-training (SST) with feature-based active learning (FAL). 
SST steps are repeated until there is little change in the probabilistic labels (less than 1\% in our experiments).
DPL learning updates the deep neural network and the graphical model parameters with warm start (i.e., the parameters are initialized with the previous parameters). 
All proposed queries are stored and won't be proposed again.
The total amount of human effort consists of generating the seed $I$ and validating $T$ queries in active learning.

S4 is a general algorithm framework that can combine various strategies for designing $\cV$, $\tt PropSST$, and $\tt PropFAL$.
In standard structure learning, $\tt PropSST$ would attempt to maximize the learning objective (e.g., conditional likelihood of seed virtual evidences) by iteratively conducting greedy structure changes. However, this is very expensive to compute, since it requires a full DPL run just to score each candidate.
Instead, we take inspiration from the feature-induction and relational learning literature and use heuristic approximations that are much faster to evaluate.
%% VE classes
In the most general setting, $\cV$ contains all possible potential functions. In practice, we can restrict it to a tractable subset to obtain a good trade-off between expressiveness and computation for the problem domain. 
Interestingly, as we will see in the experiment section, even with relatively simple classes of self-supervision, S4 can dramatically improve over DPL through structure learning and active learning.

We use text classification from natural language processing (NLP) as a running example to illustrate how to apply S4 in practice. % and make the above notions concrete.
Here, the input $X_i=(t_1,\ldots,t_{s_i})$ is a sequence of tokens and the output $Y_i$ is the classification label (e.g., $\tt pos$ or $\tt neg$ in sentiment analysis).

\subsection{Candidate self-supervision}
%\paragraph{Candidate self-supervision} 
For $\cV$, the simplest choice is to use tokens. Namely, $f_{t,l}(X_i,Y_i)=\mathbb{I}[t\in X_i\land Y_i=l]$. For simplicity, we can use a fixed weight and prior for all initial virtual evidence, i.e., $\cV=\{(f_{t,l},w,\alpha): t,l\}$. 
Take sentiment analysis as an example. $X_i$ may represent a movie review and $Y_i\in\{0,1\}$ the sentiment. 
A virtual evidence for self-supervision may stipulate that if the review contains the word ``good'', the sentiment is more likely to be positive. This can be represented by the formula $f_{\text{good},1}(X_i,Y_i) = \mathbb{I}[\text{``good''} \in X_i \wedge Y_i=1]$ with a positive weight.
A more advanced choice for $\cV$ may include high-order factors, such as $f_{ij}(Y_i,Y_j)=\mathbb{I}[Y_i=Y_j]$. If we add this factor for similar pairs $X_i, X_j$, it stipulates that instances with similar input are likely to share the same label.
Here we define similar pairs with a similarity function ${\tt Sim}(X_i,X_j)$ between $X_i$ and $X_j$, such as the cosine similarity between the sentence (or document) embeddings of $X_i$ and $X_j$, based on the current deep neural network.
Note that this is different from graph-based semi-supervised learning or other kernel-based methods in that the similarity metric is not pre-specified and fixed, but rather evolving along with the deep neural network for the end task.

\subsection{Structured self-training ({\tt\small PropSST})} 
%% propSST
From DPL learning, we obtain the current marginal estimate of the latent label variables $q_i(Y_i)$, which we would treat as probabilistic labels in assessing candidate virtual evidence. 
There are many sensible strategies for proposing candidates in structured self-training (i.e., $\tt PropSST$).
For token-based self-supervision, a common technique from the feature-selection literature is to choose a token highly correlated with a label.
For example, we can choose the token $t$ that occurs much more frequently in instances for a given label $l$ than others using our noisy label estimates.
We find that this often leads to very noisy proposals and semantic drift.
A simple refinement is to restrict our scoring to instances containing some initial self-supervised tokens.
However, this still has the drawback that a word may occur more often in instances of a class for reasons other than contributing to the label classification.
We therefore consider a more sophisticated strategy based on neural attention.
Namely, we will credit occurrences using the normalized attention weight for the given token in each instance.

Formally, let $A_{\Psi}(X_i,j)$ represent the normalized attention weight the neural network $\Psi$ assigns to the $j$-th token in $X_i$ for the final classification. We define {\em average weighted attention} for token $t$ and label $l$ as ${\tt Attn}(t,l)=\frac{1}{C_t}\sum_{i,j: X_{i,j}=t}~q_i(Y_i=l)\cdot A_{\Psi}(X_i,j)$, where $C_t$ is the number of occurrences of $t$ in $X$. 
Then $\tt PropSST$ would simply score token-based self-supervision $f_{t,l}$ using relative average weighted attention: $S_{\text{token}}(t,l)={\tt Attn}(t,l)-\sum_{l'\ne l}{\tt Attn}(t,l')$. 
In each iteration, $\tt PropSST$ picks the top scoring $f_{t,l}$ that has not been proposed yet as the new virtual evidence to add to $K$.

% entropy
We also consider an entropy-based score function that works for arbitrary input-based features. It treats the prediction module $\Psi$ as a black box, and only uses the posterior label assignments $q_i(Y_i)$. 
Consider candidate virtual evidence $f_{b,l}(X_i,Y_i) = \mathbb{I}[b(X_i) \wedge Y_i = l]$, where $b$ is a binary function over input $X_i$. This clearly generalizes token-based virtual evidence. 
Define ${\tt Ent}(b) = H\left(\frac{1}{C_b}\sum_{i: b(X_i) = 1} q_i(Y_i)\right)$,  where $H$ is the Shannon entropy and $C_b$ is the number of instances for which the feature $b$ holds true. This function represents the entropy of the average posterior among all instances with $b(X_i)=1$.
$\tt PropSST$ will then use $S_\text{entropy}(b)=1/{\tt Ent}(b)$ to choose the $b^*$ with the lowest average entropy and then pick label $l^*$ with the highest average posterior probability for $b^*$. In our experiments, this performs similarly to attention-based scores.

%% jnt-inf
For joint-inference self-supervision, we consider the similarity-based factors defined earlier, and leave the exploration of more complex factors to future work.
To distinguish task-specific similarity from pretrained similarity, we use the difference between the similarity computed using the current fine-tuned BERT model and that using the pretrained one.

Formally, let ${\tt Sim}_{\text{pretrained}}(X_i,X_j)$ be the cosine similarity between the embeddings of $X_i$ and $X_j$ generated by the pretrained BERT model, and ${\tt Sim}_{\Psi}(X_i,X_j)$ be that between the embeddings using the current learned network $\Psi$. $\tt PropSST$ would score the joint-inference factor using the relative similarity and choose the top scoring pairs to add to self-supervision:
$S_{\text{joint}}(X_i,X_j) = {\tt Sim}_{\Psi}(X_i,X_j) - {\tt Sim}_{\text{pretrained}}(X_i,X_j)$.

\subsection{Feature-based active learning ({\tt\small PropFAL})}
%% propFAL
For active learning, a common strategy is to pick the instance with highest entropy in the label distribution based on the current marginal estimate. In feature-based active learning, we can similarly pick the feature $b$ with the highest average entropy ${\tt Ent}(b)$. Note that this is opposite to how we use the entropy-based score function in $\tt PropSST$, where we choose the feature with the lowest average entropy.
In $\tt PropFAL$, we will identify $b^*=\argmax ({\tt Ent}(b))$, present $f_{b^*,l}(X,Y)=\mathbb{I}[b^*(X)\land Y=l]$ for all possible labels $l$, and ask the human expert to choose a label $l^*$ to accept or reject them all.

%% file: sections/exp.tex
\section{Experiments}
\label{sec:experiments}
\fxnote{fix captions}

%% overview: use NLP as examples; goals; eval/simulate human
We use the natural language processing (NLP) task of text classification to explore the potential for S4 to improve over DPL using structure learning and active learning. We used three standard text classification datasets: IMDb \citep{maas2011learning}, Stanford Sentiment Treebank \citep{socher2013recursive}, and Yahoo!~Answers \citep{zhang2015character}.
IMDb contains movie reviews with polarity labels (positive/negative). There are 25,000 training instances with equal numbers of positive and negative labels, and the same numbers for test.
Stanford Sentiment Treebank (StanSent) also contains movie reviews, but was annotated with five labels ranging from very negative to very positive. We used the binarized version of StanSent, which collapses the polarized categories and discards the neutral sentences. It contains 6,920 training instances and 1,821 test instances, with roughly equal split.
Overall, the StanSent reviews are shorter than IMDb's, and they often exhibit more challenging linguistic phenomena (e.g., nested negations or sarcasm).
%the oracle unigram model performs less well than on IMdB, only obtaining 79\% accuracy on the test set. On the other hand, fine-tuning a BERT-base model obtains 90\% test accuracy, so the conditional model is still able to perform well given the gold labels.
The Yahoo dataset contains 1.4 million training questions and 60,000 test questions from Yahoo!~Answers; these are equally split into 10 classes. The Yahoo results are contained in the appendix.

%% implementation details
In all our experiments with S4, we withheld gold labels from the system, used the training instances as unlabeled data, and evaluated on the test set. We reported test accuracy, as all of the datasets are class-balanced.
For our neural network prediction module $\Psi(X_i,Y_i)$, we used the standard BERT-base model pretrained using Wikipedia \citep{devlin2018bert}, along with a global-context attention layer as in \citet{yang2016hierarchical}, which we also used for attention-based scoring.
We truncated the input text to 512 tokens, the maximum allowed by the standard BERT model.
All of our baselines (except supervised bag-of-words) use the same BERT model.
For all virtual evidences, we used initial weight $w=2.2$ (the log-odds of 90\% probability) and used an $\alpha$ corresponding to an L2 penalty of $5\times10^{-8}$ on $w$. Our results are not sensitive to these values.
In all experiments, we use the Adam optimizer with an initial learning rate tuned over $[0.1, 0.01, 0.001]$. The optimizer's history is reset after each EM iteration to remove old gradient information. We always performed 3 EM iterations and trained $\Psi$ for 5 epochs per iteration.

%% VE & Simulate human supervision
For virtual evidence, we focus on token-based unary factors and similarity-based joint factors, as discussed in the previous section, and leave the exploration of more complex factors to future work. Even with these factors, our self-supervised $\Psi$ models often nearly match the accuracy of the best supervised models. %Given only token-based factors, we can also compare head-to-head
We also compare with Snorkel, a popular self-supervision system \citep{ratner2016data}. We use the latest Snorkel version  \citep{ratner2019training}, which models correlations among same-instance factors. Snorkel cannot incorporate joint-inference factors across different instances.

To simulate human supervision for unary factors, we trained a unigram model using the training data with L1 regularization and selected the 100 tokens with the highest weights for each class as the oracle self-supervision.
By default, we used the top tokens for each class in the initial self-supervision $I$. We also experimented with using random tokens from the oracle in $I$ to simulate lower-quality initial supervision and to quantify the variance of S4.
For the set of oracle joint factors, we fine-tuned the standard BERT model on the training set, used the $\tt CLS$ embedding BERT produces to compute input similarity, and picked the 100 input pairs whose similarity changed the most between the fine-tuned model and the initial model.

\begin{table}[!tb]
\fontsize{9}{11}\selectfont
\centering
        \begin{tabular}{lcc}
            \toprule
            Algorithm   & Sup. size $|I|$  & Test acc (\%) \\
            \midrule
            BoW & 25k &  87.1 \\
            DNN & 25k & 91.0  \\
            \midrule
            Self-training & 100 &  69.9   \\
            Self-training & 1k &  88.5   \\
            \midrule
            Snorkel & 6 & 76.6\\
            DPL & 6 &  80.7   \\
            S4-SST & 6     &  85.5 \\
            S4 ($T=20$) & 6 & 85.6\\
            \midrule
            Snorkel & 20 & 82.4 \\
            DPL & 20           & 78.9  \\
            S4-SST & 20     &  86.4 \\
            S4 ($T=20$) & 20 & 86.9\\
            \midrule
            Snorkel & 40 & 84.6\\
            DPL & 40 & 85.2\\
            S4-SST & 40 & 86.6 \\
            S4 ($T=20$) & 40 & 86.8\\
            \bottomrule
        \end{tabular}
      \caption{System comparison on IMDb}\label{tbl:imdb}
\end{table}

\begin{table}[h]
\fontsize{9}{11}\selectfont
      \centering
        \begin{tabular}{lcc}
            \toprule
            Algorithm & Sup. size $|I|$    & Test acc (\%) \\
            \midrule
            BoW & 6.9k & 78.9  \\
            DNN & 6.9k & 90.9 \\
            \midrule
            \multirow{2}{*}{Self-training}   & 50  & 78.0  \\
              & 100  & 81.8  \\
            \midrule
            Snorkel & 6 & 63.5\\
            DPL & 6 & 57.2\\
            S4-SST  & 6 & 73.0\\
            S4-SST + J & 6 & 76.2\\
            S4 + J ($T=20$) & 6 & 81.4\\
            \midrule
            Snorkel & 20 & 73.0\\
            DPL & 20 & 72.4\\
            S4-SST & 20 & 83.3\\
            S4-SST + J & 20 & 85.1\\
            S4 + J ($T=20$) & 20 & 84.4 \\
            \midrule
            Snorkel & 40 & 73.6\\
            DPL & 40 & 77.0\\
            S4-SST & 40 & 84.9\\
            S4-SST + J & 40 & 86.3\\
            S4 + J $(T=20)$ & 40 & 85.4\\
            \bottomrule
        \end{tabular}
        \caption{System comparison on Stanford}\label{tbl:stan}
\end{table}

\begin{table}[h]
\fontsize{9}{11}\selectfont
     \centering
        \begin{tabular}{lcccc}
            \toprule
            & $|I|=6$ & $|I|=10$ & $|I|=20$ & $|I|=40$\\
            \midrule
            $T=5$ & 79.0 & 83.8 & 84.0 & 85.9  \\
            $T=10$ & 79.5 & 83.7 & 85.0 & 86.0\\
            $T=20$ & 82.7 & 84.5 & 85.8 & 86.4\\
            \bottomrule
        \end{tabular}%
        \caption{S4-FAL results on IMDb (no SST steps)\fxnote{put these tables in the appendix and just include S4 as a row in tables 1 and 2}}\label{tbl:fal-imdb}
\end{table}
\begin{table}[h]
\fontsize{9}{11}\selectfont  
\centering
        \begin{tabular}{lcccc}
            \toprule
            & $|I|=6$ & $|I|=10$ & $|I|=20$ & $|I|=40$\\
            \midrule
            $T=5$  & 71.4  & 77.5 & 82.5 & 83.9 \\
            $T=10$ & 72.3 & 77.4 & 82.5 & 83.7 \\
            $T=20$ & 77.1 & 80.8 & 83.4 & 84.0 \\
            \bottomrule
        \end{tabular}
        \caption{S4-FAL results on Stanford (no SST steps)}\label{tbl:fal-stan}
\end{table}

%% structure learning
We first investigate whether structure learning can help in S4 by running without feature-based active learning. We set the query budget $T=0$ in Algorithm \ref{alg:S4}. Because we only take structured self-training steps when $T=0$, we denote this version of S4 as S4-SST.
Table~\ref{tbl:imdb} shows the results on IMDb.
With just six self-supervised tokens (three per class), S4-SST already attained 86\% test accuracy, which outperforms self-training with 100 labeled examples by 16 absolute points, and is only slightly worse than self-training with 1000 labeled examples or supervised training with 25,000 labeled examples.
By conducting structure learning, S4-SST substantially outperformed DPL, gaining about 5 absolute points in accuracy (a 25\% relative reduction in error), and also outperformed the Snorkel baseline by 8.9 points.
Interestingly, with more self-supervision at 20 tokens, DPL's performance drops slightly, which might stem from more noise in the initial self-supervision. By contrast, S4-SST capitalized on the larger seed self-supervision and attained steady improvement.
Even with substantially more self-supervision at 40 tokens, S4-SST still attained similar accuracy gain, demonstrating the power in structure learning.
On average across different initial amounts of supervision $|I|$, S4-SST outperforms DPL by 5.6 points and Snorkel by 5.3 points.
Next, we consider the full S4 algorithm with a budget of up to $T=20$ human queries (S4 ($T=20$)). Overall, by automatically generating self-supervision from structure learning, S4-SST already attained very high accuracy on this dataset. However, active learning can still produce some additional gain.
The only randomness in Table \ref{tbl:imdb} is the initialization of the deep network $\Psi$, which has a negligible effect.

\begin{figure*}[t]
\centering
\begin{subfigure}{0.25\linewidth}
\includegraphics[width=\linewidth]{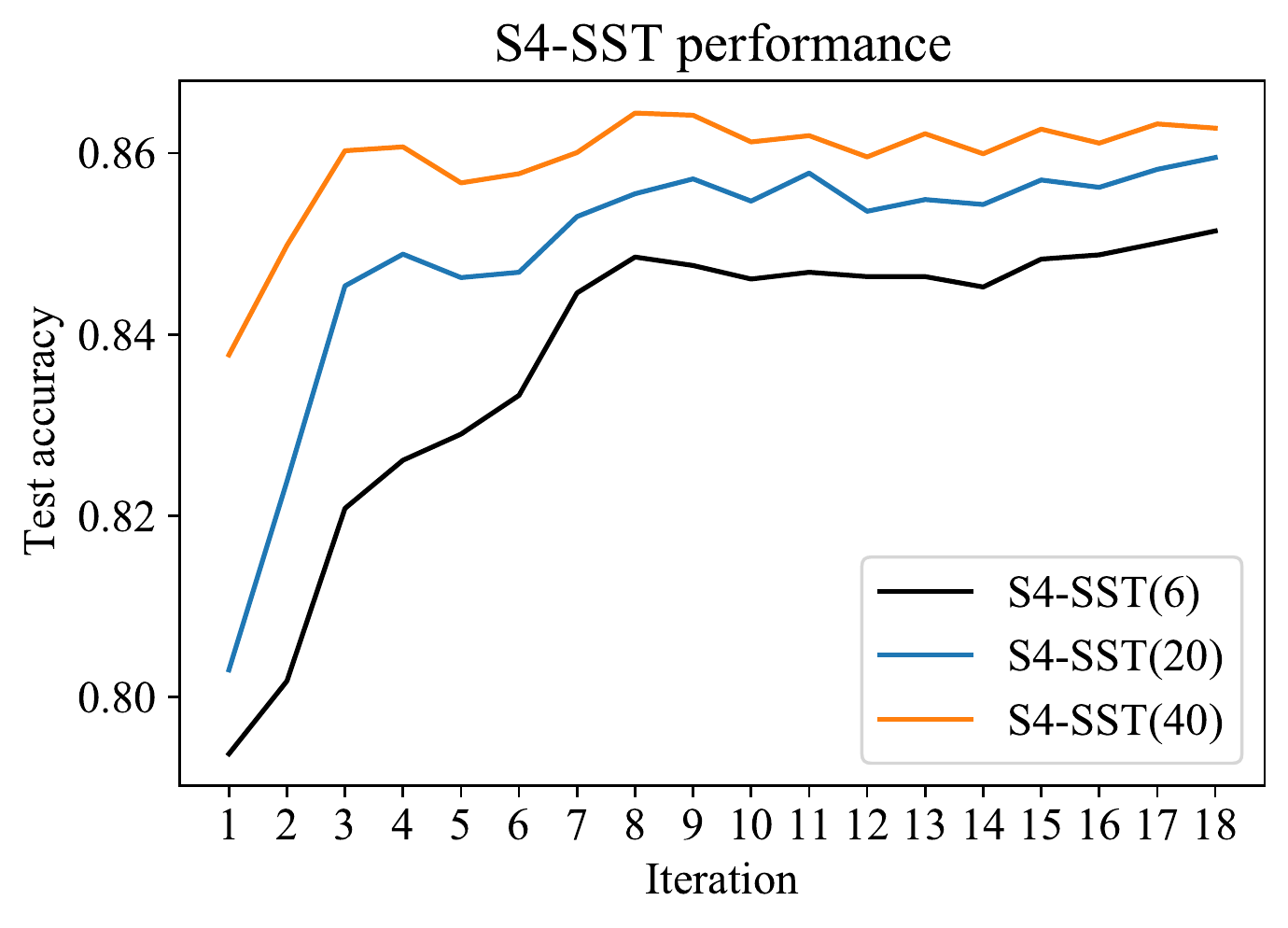}
\end{subfigure}%
\begin{subfigure}{0.25\linewidth}
\includegraphics[width=\linewidth]{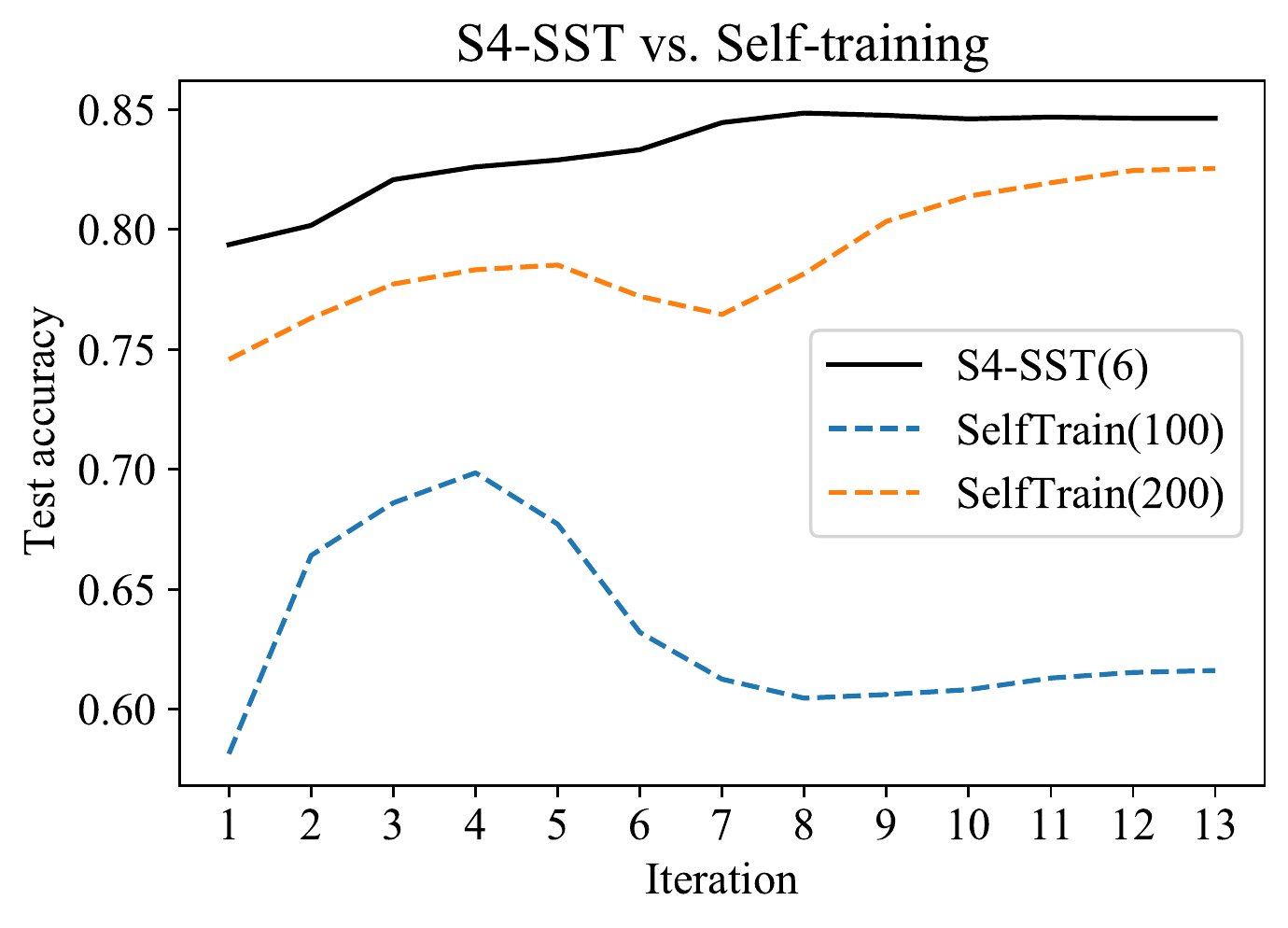}
\end{subfigure}%
\begin{subfigure}{0.25\linewidth}
\includegraphics[width=\linewidth]{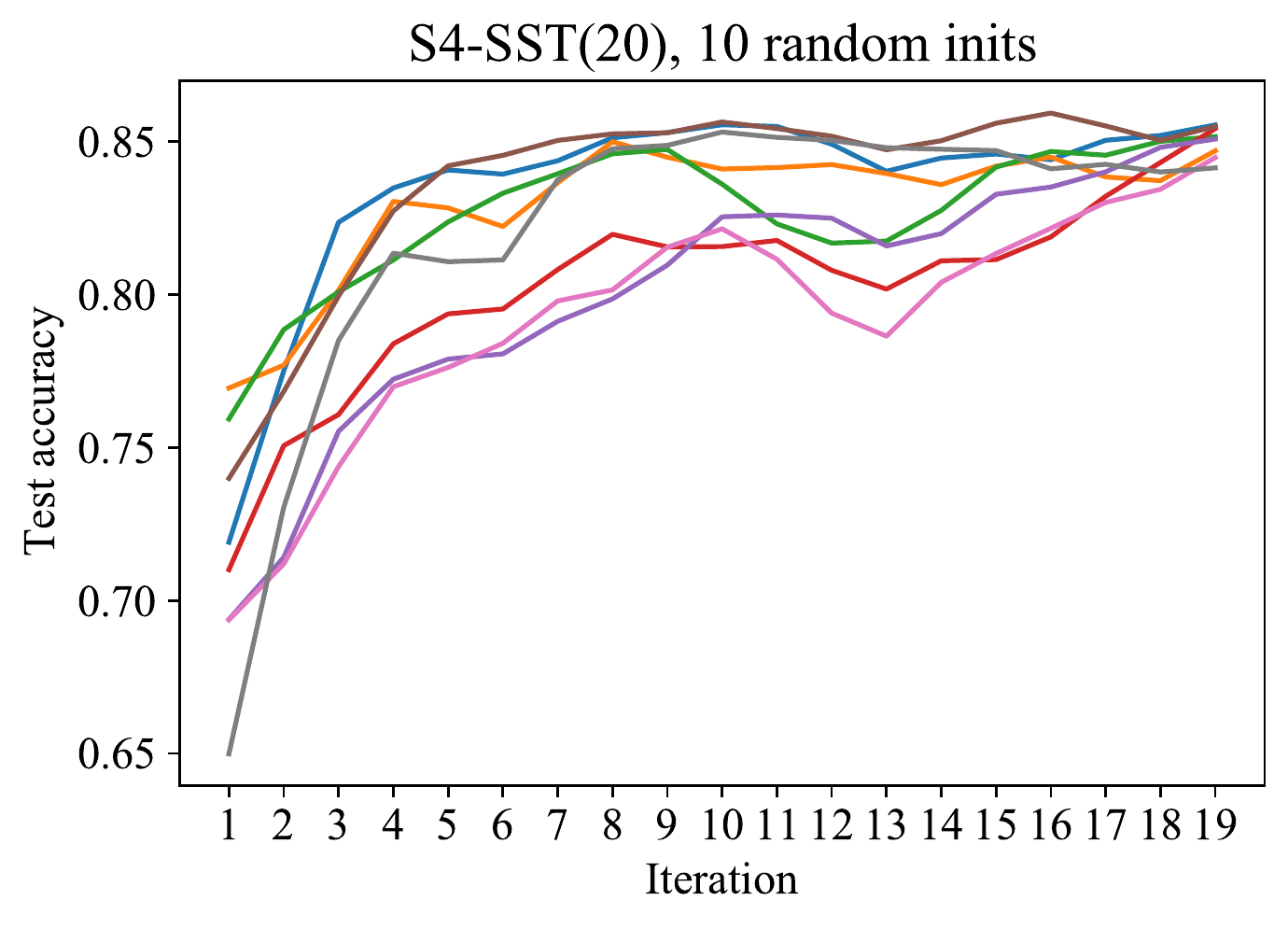}
\end{subfigure}%
\begin{subfigure}{0.25\linewidth}
  \centering
  \resizebox{\linewidth}{!}{%
    \begin{tabular}{cc|cc}
        \toprule
        Init. pos.      & Init. neg. & New pos. & New neg. \\
        \midrule
        refreshing &  lacks & superb &  terrible \\
        moving     &  waste & lovely     &  horrible \\
        caught     &  bland & touching    &  negative \\
        magic      &  stink & charming      &  boring \\
        captures   &   1    & delightful   &   2    \\
        \bottomrule
    \end{tabular}}
\end{subfigure}
\caption{S4-SST learning in IMDb.
Left-to-right: S4-SST learning curves with various numbers of initial self-supervised tokens;
comparison between the learning curves of S4-SST with just six self-supervised tokens and self-training with an order of magnitude more direct supervision (100/200 labeled examples); %Note the different $y$-axis scales.
%S4-SST obtains 86\% test accuracy with 10 initial factors per class (20 total), whereas self-training requires 1000 initial labels to reach approximately the same accuracy.
multiple runs of S4-SST with 20 \emph{random} oracle self-supervised tokens (rather than the top oracle tokens); pre-specified self-supervised tokens and proposed tokens in the first few iterations.
}
\label{fig:sst-best}
\end{figure*}

Figure~\ref{fig:sst-best} (leftmost) shows how S4-SST iterations improve the test accuracy of the learned neural network with different amounts of initial virtual evidence.
Not surprisingly, with more initial self-supervision, the gain is less pronounced, but still significant.
Figure~\ref{fig:sst-best} (center left) compares the learning curves of S4-SST with those of self-training.
Remarkably, with just \emph{six} self-supervised tokens, S4-SST not only attained substantial gain over the iterations, but also easily outperformed self-training despite the latter using an order of magnitude more label information (up to 200 labeled examples).
This shows that S4 is much more effective in leveraging bounded human effort for supervision.

%%%%%%%%%%%%%%% w/o supervised attn: 1% in avg // skip

Figure~\ref{fig:sst-best} (center right) shows 10 runs of S4-SST when it was initialized with 20 \emph{random} oracle tokens (10 per class), rather than the \emph{top} 20 tokens from the oracle. As expected, DPL's initial performance is worse than with the top oracle tokens.
However, over the iterations, S4-SST was able to recover even from particularly poor initial state, gaining up to 20 absolute accuracy points over DPL in the process.
The final accuracy of S4-SST is 85.2 $\pm$ 0.9, compared to 71.5 $\pm$ 6.5 for DPL and 72.9 $\pm$ 6.7 for Snorkel, a mean improvement of more than 12 absolute accuracy points over both baselines.
S4-SST's gains over DPL and Snorkel are statistically significant using a paired $t$-test (samples are paired when the algorithms have the same initial factors $I$) with $p=0.01$.
This indicates that S4-SST is robust to noise in the initial self-supervision.
Figure~\ref{fig:sst-best} (rightmost) shows the initial self-supervised tokens and the ones learned by S4-SST in the first few iterations. We can see that S4-SST is able to learn highly relevant self-supervision tokens.

S4 has similar gains over DPL, Snorkel, and self-training on the Stanford dataset. See Table~\ref{tbl:stan}.
The Stanford dataset is much more challenging and for a long while, it was hard to exceed 80\% test accuracy \citep{socher2013recursive}. 
Interestingly, S4-SST was able to surpass this milestone using just 20 initial self-supervision tokens. As in Table \ref{tbl:imdb}, the only randomness is in the initialization of $\Psi$, which is negligible.

For IMDb, as can be seen above, even with simple token-based self-supervision, S4-SST already performed extremely well. So we focused our investigation of joint-inference factors on the more challenging Stanford dataset (S4-SST + J).
While S4-SST already performed very well with token-based self-supervision, by incorporating joint-inference factors, it could still gain up to 3 absolute accuracy points.
An example learned joint-inference factor is between the sentence pair:
{\em This is no ``Waterboy!"} and {\em ``It manages to accomplish what few sequels can---it equals the original and in some ways even better"}.
Note that Waterboy is widely considered a bad movie, hence the first sentence expresses a positive sentiment, just like the second.
It is remarkable that S4 can automatically induce such factors with complex and subtle semantics from small initial self-supervision.

The Stanford results also demonstrated that active learning could play a bigger role in more challenging scenarios. With limited initial self-supervision ($|I|=6$), the full S4 system (S4+J (T=20)) gained 8 absolute points over S4-SST and 5 absolute points over S4-SST+J. With sufficient initial self-supervision and joint-inference, however, active learning was actually slightly detrimental ($|I|=20,40$).

Finally, we evaluate S4-FAL, which conducts active learning but not structure learning. See Tables~\ref{tbl:fal-imdb} and \ref{tbl:fal-stan}. As expected, performance improved with larger initial self-supervision ($I$) and human query budget ($T$). Active learning helps the most when initial self-supervision is limited. Compared to S4 with structure learning, however, active learning alone is less effective. For example, without requiring any human queries, S4-SST outperformed S4-FAL on both IMDB and Stanford even when the latter was allowed up to $T=20$ human queries.

%% file: sections/related.tex
\section{Related work}

%\subsection{Self supervision}
Techniques to compensate for the lack of direct supervision come in many names and forms \citep{mintz2009distant, ratner2016data,bach2017learning,roth2017incidental,wang2018deep}.
Self-supervision has emerged as an encompassing paradigm that views these as instances of using self-specified templates to generate noisy labeled examples on unlabeled data.
The name {\em self-supervision} is closely related to {\em self-training} \cite{mcclosky2006effective}, which bootstraps from a supervised classifier, uses it to annotate unlabeled instances, and iteratively uses the confident labels to retrain the classifier. 
{\em Task-agnostic self-supervision} generalizes word embedding and language modeling by learning to predict self-specified masked tokens, as exemplified by recent pretraining methods such as BERT \citep{devlin2018bert}.
In this paper, we focus on {\em task-specific self-supervision} and use pretrained models as a building block for task-specific learning.

%% DP - generalize earlier anchor learning / per instance
Existing self-supervision paradigms are typically special cases of deep probabilistic logic (DPL). E.g., the popular {\em data programming} methods \citep{ratner2016data, bach2017learning,varma2017inferring} admit only virtual evidences for individual instances (labeling functions or their correlations). 
{\em Anchor learning} \citep{halpern2016electronic} is an earlier form of data programming that, while more restricted, allows for stronger theoretical learning guarantees. %assumes conditional independence among the labeling functions (anchors).
%% Jnt: PR/GE, Prototype special ex
{\em Prototype learning} is an even earlier special case with labeling functions provided by ``prototypes'' \citep{haghighi2006prototype, poon2013grounded}.
Using Markov logic to model self-supervision, DPL can incorporate arbitrary prior beliefs on both individual labels and their interdependencies, thereby unleashing the full power of {\em joint inference} \citep{chang2007guiding, druck2008learning, poon2008joint, ganchev2010posterior} to amplify and propagate self-supervision signals.

%% structure learning sup / bootstrap - inducing new rules / snuba
Self-supervised self-supervision (S4) further extends DPL with structure learning capability. Most structure learning techniques are developed for the supervised setting, where structure search is guided by labeled examples \citep{koller-struc-lrn,kok2005learning}. 
Moreover, traditional relational learning induces deterministic rules and is susceptible to noise and uncertainty.
{\em Bootstrapping} learning is one of the earliest and simplest self-supervision methods with some rule-learning capability, by alternating between inducing characteristic contextual patterns and classifying instances \cite{hearst1992automatic, NELL}. The pattern classes are limited and only applicable to special problems (e.g., ``A such as B" to find $\tt ISA$ relations). Most importantly, they lack a coherent probabilistic formulation and may suffer catastrophic {\em semantic drift} due to ambiguous patterns (e.g., ``cookie'' as food or compute use). %Unsupervised rule bootstrapping goes back at least to \citet{yarowsky1995unsupervised} and \citet{collins1999unsupervised}. S4 combines these ideas with DPL and prediction modules that consist of deep neural networks.
\citet{yarowsky1995unsupervised} and \citet{collins1999unsupervised} designed a more sophisticated rule induction approach, but their method uses deterministic rules and may be sensitive to noise and ambiguity.
Recently, Snuba \citep{varma2018snuba} extends the data programming framework by automatically adding new labeling functions, but like prior data programming methods, their self-supervision framework is limited to modeling prior beliefs on individual instances. Their method also requires access to a small number of labeled examples to score new labeling functions.
\fxnote{what about boosting rules? and socratic?}

%% active learning
Another significant advance in S4 is by extending DPL with the capability to conduct {\em structured active learning}, where human experts are asked to verify arbitrary virtual evidences, rather than a label decision.
Note that by admitting joint inference factors, this is more general than prior use of {\em feature-based active learning}, which focuses on per-instance features \citep{druck2009active}.
As our experiments show, interleaving structured self-training learning and structured active learning results in substantial gains, and provides the best use of precious human bandwidth. \citet{tong2001active} previously considered active structure learning in the context of Bayesian networks.
Anchor learning \citep{halpern2016electronic} can also suggest new self-supervision for human review.
Darwin \citep{galhotra2020adaptive} incorporates active learning for verifying proposed rules, but it doesn't conduct structure learning, and like Snuba and other data programming methods, it only models individual instances.

%% combine DL w. GM
Neural-symbolic learning and reasoning has received increasing attention \cite{besold&al17}. In particular, combining deep learning with probabilistic models can leverage their complementary strengths in modeling complex patterns and infusing rich prior knowledge.
Prior work tends to focus on deep generative models that aim to uncover latent factors for generative modeling and semi-supervised learning \citep{kingma2013auto,kingma2014semi}.
They admit limited forms of self-supervision (e.g., latent structures such as Markov chains \citep{johnson2016composing}).
%{\em Deep probabilistic programming} provides a flexible interface for exploring such composition \cite{edward}.
S4 and DPL instead combine a discriminative neural network predictor with a generative self-supervision model based on Markov logic, and can fully leverage their respective capabilities to advance co-learning \citep{blum1998combining,grechkin2017ezlearn}.
Deep neural networks also provide a powerful feature-induction engine to support structure learning and active learning.

%% file: sections/appendix.tex
\onecolumn
\section{Supplementary material}

\subsection{Structured self-training convergence}
In Algorithm \ref{alg:S4}, structured self-training (SST) iterations are repeated until convergence in the while loop. Here we elaborate on the convergence criterion described in the main text (Self-Supervised Self-Supervision Section). Intuitively, convergence occurs when the expected latent labels change little despite the addition of new self-supervision from SST.

Formally, consider the set $\Delta = \{i  | \argmax_{y_i} E_{\Phi^{t-1}(X,Y)}[y_i] \ne \argmax_{y_i} E_{\Phi^{t}(X,Y)}[y_i]\}$. This is the set of instances for which the labels based on self-supervision alone (\emph{excluding} the neural network prediction module $\Psi$) have changed between subsequent iterations. We stop the SST iterations once $|\Delta|/N < \alpha$ for some small $\alpha$, as we don't expect there will be much change afterwards. We used $\alpha=1\%$, which worked well in preliminary experiments, and performed no further tuning.
%Note that ties in the virtual evidence posterior are broken deterministically in a consistent manner across iterations. Once this criterion fires, S4 switches to a feature-based active learning query.

\subsection{Count normalization in score functions}
All the score functions in $\tt PropSST$ and $\tt PropFAL$ normalize the score using the feature count (e.g., $C_t$, $C_b$). In practice, if we apply this to all features, we may inadvertently promote rare features. Thus \citet{druck2009active} additionally multiplied by the logarithm of the count (i.e., they normalize using $\frac{\log C_t}{C_t}$). By contrast, we found it preferable to use standard count normalization, but simply skip the rare features. In all of our experiments, we only consider the top 2.5\% most frequent features. We found that this worked well in preliminary experiments on the held-out data in IMDb, and our results are not sensitive to this value.

\subsection{Additional experimental details and results}
\subsubsection{Yahoo.}
We provide additional results for S4-SST on Yahoo using token-based factors and attention scoring. See Table \ref{tbl:yahoo}. We focused our experiments on a fixed 10\% of the training set due to its very large size (1.4 million examples). 
There are 10 classes, so initial self-supervision size of 50 (100) represents 5 (10) initial tokens per class (for S4, DPL, and Snorkel), or 5 (10) labeled examples per class (for self-training). 
Compared to binary sentiment analysis in IMDb and Stanford, Yahoo represents a much more challenging dataset, with ten classes and larger input text for each instance. The linguistic phenomena are much more diverse, and therefore neither Snorkel or DPL performed much better than self-training, as a token-based self-supervision confers not much more information than a labeled example. However, S4-SST is still able to attain substantial improvement over the initial self-supervision. E.g., with initial supervision size of 100, S4-SST gained about 11 absolute accuracy points over DPL, and 16 absolute points over Snorkel.  
Additionally, S4-SST is able to better utilize the new factors than Snorkel. If we run Snorkel using the same initial factors as S4-SST and also add the new factors proposed by S4-SST in each iteration, the accuracy improved from 36.5 to 44.2, but still trailed S4-SST (52.3) by 8 absolute points. 
Interestingly, both DPL and Snorkel perform better on Yahoo with \emph{fewer} initial factors, at 5 per class, suggesting they are sensitive to noise in the less reliable initial self-supervision. By contrast, S4-SST is more noise-tolerant and benefits from additional initial supervision.

\begin{table}[tb]
      \centering
        \begin{tabular}{lcc}
            \toprule
            Algorithm   & Sup. size   & Test acc (\%) \\
            \midrule
            BoW & 140k &  71.2   \\
            DNN & 140k &  79.8  \\
            \midrule
            Self-training & 50 & 38.4 \\
            Snorkel & 50  & 37.2 \\        
            DPL & 50  & 41.8  \\
            S4-SST & 50 & 49.1  \\            
            \midrule
            Self-training & 100 & 38.2  \\
            Snorkel & 100  & 36.5\\        
            DPL & 100  & 41.7  \\
            S4-SST & 100 & 52.3  \\
            \bottomrule
        \end{tabular}
        \caption{Comparison of test accuracy on Yahoo.}\label{tbl:yahoo}
\end{table}
\begin{table}[tb]
\centering
    \begin{subfigure}{.4\linewidth}
        \centering
        \begin{tabular}{ccc}
        %\centering
            \toprule
            $|I|$ & Entropy-based  & Attention-based\\
            \midrule
            6 & 82.1 & 85.5 \\
            20 & 84.9 & 86.4 \\
            40 & 85.5 & 86.6\\
            \bottomrule
        \end{tabular}
        \caption{IMDb}\label{tbl:ent-imdb}
    \end{subfigure}%
    \begin{subfigure}{.4\linewidth}
        \centering
        \begin{tabular}{ccc}
         %\centering
            \toprule
            $|I|$ & Entropy-based  & Attention-based\\
            \midrule
            6 & 77.2 & 73.0\\
            20 & 85.1 & 83.3 \\
            40 & 85.7 & 84.9 \\
            \bottomrule
        \end{tabular}
        \caption{Stanford}\label{tbl:ent-stan}
    \end{subfigure}
    \caption{Comparison of test accuracy on IMDb and Stanford. Entropy-based scoring ($S_\text{entropy}$) perform comparably as attention-based scoring ($S_\text{token}$).}
\end{table}

\subsubsection{Entropy-based scoring.}\label{sec:alt-scores}
As stated in the main text, entropy-based scoring ($S_\text{entropy}$) is a more general scoring function that works for arbitrary features in self-supervision. 
Tables \ref{tbl:ent-imdb} and \ref{tbl:ent-stan} show the results comparing S4-SST test accuracy using entropy-based scoring ($S_\text{entropy}$) and attention-based scoring ($S_\text{token}$). Entropy-based scoring slightly outperforms attention-based scoring on Stanford Sentiment and slightly trails on IMDb. Overall, the two perform comparably but entropy-based scoring is more generally applicable.

\subsubsection{Hyperparameters}
 In all of our experiments, we used three variational EM iterations and trained the deep neural network for 5 epochs per EM iteration. For the global-context attention layer, we used a context dimension of 5. The model is warm-started across EM iterations (in DPL), but \emph{not} across the outer iterations in S4 (the for loop). In all experiments, we used the Adam optimizer with an initial learning rate tuned over $[0.1, 0.01, 0.001]$. The optimizer's history is reset after each EM iteration to remove old gradient information.
 In all of our Snorkel baselines, we separately tuned the initial learning rate over the same set, and trained the deep neural network for the same number of {\em total} epochs that DPL uses to ensure a fair comparison.

%% file: s4-arxiv.bbl
\begin{thebibliography}{}

\bibitem[Bach et~al., 2017]{bach2017learning}
Bach, S.~H., He, B., Ratner, A., and R{\'e}, C. (2017).
\newblock Learning the structure of generative models without labeled data.
\newblock In {\em Proceedings of the 34th International Conference on Machine
  Learning-Volume 70}, pages 273--282.

\bibitem[Besold et~al., 2017]{besold&al17}
Besold, T.~R., d'Avila Garcez, A., Bader, S., Bowman, H., Domingos, P.,
  Hitzler, P., Kühnberger, K.-U., Lamb, L.~C., Lowd, D., Lima, P. M.~V.,
  de~Penning, L., Pinkas, G., Poon, H., and Zaverucha, G. (2017).
\newblock Neural-symbolic learning and reasoning: A survey and interpretation.
\newblock In {\em arxiv}.

\bibitem[Blum and Mitchell, 1998]{blum1998combining}
Blum, A. and Mitchell, T. (1998).
\newblock Combining labeled and unlabeled data with co-training.
\newblock In {\em Proceedings of the eleventh annual conference on
  Computational learning theory}, pages 92--100.

\bibitem[Carlson et~al., 2010]{NELL}
Carlson, A., Betteridge, J., Kisiel, B., Settles, B., Hruschka, E.~R., and
  Mitchell, T.~M. (2010).
\newblock Toward an architecture for never-ending language learning.
\newblock In {\em Twenty-Fourth AAAI Conference on Artificial Intelligence}.

\bibitem[Chang et~al., 2007]{chang2007guiding}
Chang, M.-W., Ratinov, L., and Roth, D. (2007).
\newblock Guiding semi-supervision with constraint-driven learning.
\newblock In {\em Proceedings of the 45th annual meeting of the association of
  computational linguistics}, pages 280--287.

\bibitem[Collins and Singer, 1999]{collins1999unsupervised}
Collins, M. and Singer, Y. (1999).
\newblock Unsupervised models for named entity classification.
\newblock In {\em 1999 Joint SIGDAT Conference on Empirical Methods in Natural
  Language Processing and Very Large Corpora}.

\bibitem[Devlin et~al., 2018]{devlin2018bert}
Devlin, J., Chang, M.-W., Lee, K., and Toutanova, K. (2018).
\newblock Bert: Pre-training of deep bidirectional transformers for language
  understanding.
\newblock {\em arXiv preprint arXiv:1810.04805}.

\bibitem[Domke, 2013]{domke2013learning}
Domke, J. (2013).
\newblock Learning graphical model parameters with approximate marginal
  inference.
\newblock {\em IEEE transactions on pattern analysis and machine intelligence},
  35(10):2454--2467.

\bibitem[Druck et~al., 2008]{druck2008learning}
Druck, G., Mann, G., and McCallum, A. (2008).
\newblock Learning from labeled features using generalized expectation
  criteria.
\newblock In {\em Proceedings of the 31st annual international ACM SIGIR
  conference on Research and development in information retrieval}, pages
  595--602.

\bibitem[Druck et~al., 2009]{druck2009active}
Druck, G., Settles, B., and McCallum, A. (2009).
\newblock Active learning by labeling features.
\newblock In {\em Proceedings of the 2009 Conference on Empirical Methods in
  Natural Language Processing: Volume 1-Volume 1}, pages 81--90. Association
  for Computational Linguistics.

\bibitem[Friedman and Koller, 2003]{koller-struc-lrn}
Friedman, N. and Koller, D. (2003).
\newblock Being bayesian about network structure. a bayesian approach to
  structure discovery in bayesian networks.
\newblock {\em Machine learning}, 50(1-2):95--125.

\bibitem[Galhotra et~al., 2020]{galhotra2020adaptive}
Galhotra, S., Golshan, B., and Tan, W.-C. (2020).
\newblock Adaptive rule discovery for labeling text data.
\newblock {\em arXiv preprint arXiv:2005.06133}.

\bibitem[Ganchev et~al., 2010]{ganchev2010posterior}
Ganchev, K., Gra{\c{c}}a, J., Gillenwater, J., and Taskar, B. (2010).
\newblock Posterior regularization for structured latent variable models.
\newblock {\em Journal of Machine Learning Research}, 11(67):2001--2049.

\bibitem[Grechkin et~al., 2017]{grechkin2017ezlearn}
Grechkin, M., Poon, H., and Howe, B. (2017).
\newblock Ezlearn: Exploiting organic supervision in large-scale data
  annotation.
\newblock {\em arXiv preprint arXiv:1709.08600}.

\bibitem[Haghighi and Klein, 2006]{haghighi2006prototype}
Haghighi, A. and Klein, D. (2006).
\newblock Prototype-driven learning for sequence models.
\newblock In {\em Proceedings of the main conference on Human Language
  Technology Conference of the North American Chapter of the Association of
  Computational Linguistics}, pages 320--327. Association for Computational
  Linguistics.

\bibitem[Hall, 1999]{feat-sel}
Hall, M.~A. (1999).
\newblock Correlation-based feature selection for machine learning.

\bibitem[Halpern et~al., 2016]{halpern2016electronic}
Halpern, Y., Horng, S., Choi, Y., and Sontag, D. (2016).
\newblock Electronic medical record phenotyping using the anchor and learn
  framework.
\newblock {\em Journal of the American Medical Informatics Association},
  23(4):731--740.

\bibitem[Hearst, 1992]{hearst1992automatic}
Hearst, M.~A. (1992).
\newblock Automatic acquisition of hyponyms from large text corpora.
\newblock In {\em Proceedings of the 14th conference on Computational
  linguistics-Volume 2}, pages 539--545. Association for Computational
  Linguistics.

\bibitem[Johnson et~al., 2016]{johnson2016composing}
Johnson, M.~J., Duvenaud, D.~K., Wiltschko, A., Adams, R.~P., and Datta, S.~R.
  (2016).
\newblock Composing graphical models with neural networks for structured
  representations and fast inference.
\newblock In {\em Advances in neural information processing systems}, pages
  2946--2954.

\bibitem[Kingma et~al., 2014]{kingma2014semi}
Kingma, D.~P., Mohamed, S., Rezende, D.~J., and Welling, M. (2014).
\newblock Semi-supervised learning with deep generative models.
\newblock In {\em Advances in neural information processing systems}, pages
  3581--3589.

\bibitem[Kingma and Welling, 2013]{kingma2013auto}
Kingma, D.~P. and Welling, M. (2013).
\newblock Auto-encoding variational bayes.
\newblock {\em arXiv preprint arXiv:1312.6114}.

\bibitem[Kok and Domingos, 2005]{kok2005learning}
Kok, S. and Domingos, P. (2005).
\newblock Learning the structure of markov logic networks.
\newblock In {\em Proceedings of the 22nd international conference on Machine
  learning}, pages 441--448.

\bibitem[LeCun et~al., 2015]{lecun2015deep}
LeCun, Y., Bengio, Y., and Hinton, G. (2015).
\newblock Deep learning.
\newblock {\em Nature}, 521(7553):436--444.

\bibitem[Maas et~al., 2011]{maas2011learning}
Maas, A.~L., Daly, R.~E., Pham, P.~T., Huang, D., Ng, A.~Y., and Potts, C.
  (2011).
\newblock Learning word vectors for sentiment analysis.
\newblock In {\em Proceedings of the 49th annual meeting of the association for
  computational linguistics: Human language technologies-volume 1}, pages
  142--150. Association for Computational Linguistics.

\bibitem[McClosky et~al., 2006]{mcclosky2006effective}
McClosky, D., Charniak, E., and Johnson, M. (2006).
\newblock Effective self-training for parsing.
\newblock In {\em Proceedings of the main conference on human language
  technology conference of the North American Chapter of the Association of
  Computational Linguistics}, pages 152--159. Association for Computational
  Linguistics.

\bibitem[Mintz et~al., 2009]{mintz2009distant}
Mintz, M., Bills, S., Snow, R., and Jurafsky, D. (2009).
\newblock Distant supervision for relation extraction without labeled data.
\newblock In {\em Proceedings of the Joint Conference of the 47th Annual
  Meeting of the ACL and the 4th International Joint Conference on Natural
  Language Processing of the AFNLP: Volume 2-Volume 2}, pages 1003--1011.
  Association for Computational Linguistics.

\bibitem[Murphy et~al., 1999]{murphy1999loopy}
Murphy, K.~P., Weiss, Y., and Jordan, M.~I. (1999).
\newblock Loopy belief propagation for approximate inference: an empirical
  study.
\newblock In {\em Proceedings of the Fifteenth conference on Uncertainty in
  artificial intelligence}, pages 467--475.

\bibitem[Pearl, 1988]{pearl1988probabilistic}
Pearl, J. (1988).
\newblock {\em Probabilistic Reasoning in Intelligent Systems: Networks of
  Plausible Inference}.
\newblock Morgan Kaufmann.

\bibitem[Poon, 2013]{poon2013grounded}
Poon, H. (2013).
\newblock Grounded unsupervised semantic parsing.
\newblock In {\em Proceedings of the 51st Annual Meeting of the Association for
  Computational Linguistics (Volume 1: Long Papers)}, pages 933--943.

\bibitem[Poon and Domingos, 2008]{poon2008joint}
Poon, H. and Domingos, P. (2008).
\newblock Joint unsupervised coreference resolution with markov logic.
\newblock In {\em Proceedings of the 2008 conference on empirical methods in
  natural language processing}, pages 650--659.

\bibitem[Ratner et~al., 2019]{ratner2019training}
Ratner, A., Hancock, B., Dunnmon, J., Sala, F., Pandey, S., and R{\'e}, C.
  (2019).
\newblock Training complex models with multi-task weak supervision.
\newblock In {\em Proceedings of the AAAI Conference on Artificial
  Intelligence}, volume~33, pages 4763--4771.

\bibitem[Ratner et~al., 2016]{ratner2016data}
Ratner, A.~J., De~Sa, C.~M., Wu, S., Selsam, D., and R{\'e}, C. (2016).
\newblock Data programming: Creating large training sets, quickly.
\newblock In {\em Advances in neural information processing systems}, pages
  3567--3575.

\bibitem[Richardson and Domingos, 2006]{richardson&domingos06}
Richardson, M. and Domingos, P. (2006).
\newblock Markov logic networks.
\newblock {\em Machine learning}, 62(1-2):107--136.

\bibitem[Roth, 2017]{roth2017incidental}
Roth, D. (2017).
\newblock Incidental supervision: Moving beyond supervised learning.
\newblock In {\em Thirty-First AAAI Conference on Artificial Intelligence}.

\bibitem[Socher et~al., 2013]{socher2013recursive}
Socher, R., Perelygin, A., Wu, J., Chuang, J., Manning, C.~D., Ng, A.~Y., and
  Potts, C. (2013).
\newblock Recursive deep models for semantic compositionality over a sentiment
  treebank.
\newblock In {\em Proceedings of the 2013 conference on empirical methods in
  natural language processing}, pages 1631--1642.

\bibitem[Tong and Koller, 2001]{tong2001active}
Tong, S. and Koller, D. (2001).
\newblock Active learning for structure in bayesian networks.
\newblock In {\em International joint conference on artificial intelligence},
  volume~17, pages 863--869. Citeseer.

\bibitem[Varma et~al., 2017]{varma2017inferring}
Varma, P., He, B.~D., Bajaj, P., Khandwala, N., Banerjee, I., Rubin, D., and
  R{\'e}, C. (2017).
\newblock Inferring generative model structure with static analysis.
\newblock In {\em Advances in neural information processing systems}, pages
  240--250.

\bibitem[Varma and R{\'e}, 2018]{varma2018snuba}
Varma, P. and R{\'e}, C. (2018).
\newblock Snuba: automating weak supervision to label training data.
\newblock {\em Proceedings of the VLDB Endowment}, 12(3):223--236.

\bibitem[Verbeek and Triggs, 2008]{verbeek2008scene}
Verbeek, J. and Triggs, B. (2008).
\newblock Scene segmentation with conditional random fields learned from
  partially labeled images.
\newblock In {\em Proc. NIPS}.

\bibitem[Wang and Poon, 2018]{wang2018deep}
Wang, H. and Poon, H. (2018).
\newblock Deep probabilistic logic: A unifying framework for indirect
  supervision.
\newblock In {\em Proceedings of the 2018 Conference on Empirical Methods in
  Natural Language Processing}, pages 1891--1902.

\bibitem[Yang et~al., 2016]{yang2016hierarchical}
Yang, Z., Yang, D., Dyer, C., He, X., Smola, A., and Hovy, E. (2016).
\newblock Hierarchical attention networks for document classification.
\newblock In {\em Proceedings of the 2016 conference of the North American
  chapter of the association for computational linguistics: human language
  technologies}, pages 1480--1489.

\bibitem[Yarowsky, 1995]{yarowsky1995unsupervised}
Yarowsky, D. (1995).
\newblock Unsupervised word sense disambiguation rivaling supervised methods.
\newblock In {\em 33rd annual meeting of the association for computational
  linguistics}, pages 189--196.

\bibitem[Zhang et~al., 2015]{zhang2015character}
Zhang, X., Zhao, J., and LeCun, Y. (2015).
\newblock Character-level convolutional networks for text classification.
\newblock In {\em Advances in neural information processing systems}, pages
  649--657.

\end{thebibliography}
